\documentclass{article}

\usepackage{PRIMEarxiv}

\usepackage[utf8]{inputenc} %
\usepackage[T1]{fontenc}    %
\usepackage{hyperref}       %
\usepackage{url}            %
\usepackage{booktabs}       %
\usepackage{amsfonts}       %
\usepackage{nicefrac}       %
\usepackage{microtype}      %
\usepackage{lipsum}
\usepackage{fancyhdr}       %
\usepackage{graphicx}       %
\graphicspath{{media/}}     %
\usepackage{multirow}
\usepackage{caption}
\usepackage{subcaption}

\title{CoLI-Machine Learning Approaches for Code-mixed Language Identification at the Word Level in Kannada-English Texts}

\author{
  H.L. Shashirekha \textsuperscript{\rm 1, a}, F. Balouchzahi\textsuperscript{\rm 2, b}, M.D. Anusha\textsuperscript{\rm 1, c}, G. Sidorov\textsuperscript{\rm 2, d} \\
 \textsuperscript{\rm 1}Department of Computer Science, Mangalore University, Mangalore, India \\
  \textsuperscript{\rm 2}Instituto Politécnico Nacional (IPN), Center for Computing Research (CIC), Mexico\\
  \texttt{\{\textsuperscript{\rm a}hlsrekha, \textsuperscript{\rm c}anusha\}@gmail.com,\{\textsuperscript{\rm b}fbalouchzahi2021, \textsuperscript{\rm d}sidorov\}@cic.ipn.mx} \\
}

\begin{document}
\maketitle

\begin{abstract}
The task of automatically identifying a language used in a given text is called Language Identification (LI). India is a multilingual country and many Indians especially youths are comfortable with Hindi and English, in addition to their local languages. Hence, they often use more than one language to post their comments on social media. Texts containing more than one language are called “code-mixed texts” and are a good source of input for LI. Languages in these texts may be mixed at sentence level, word level or even at sub-word level. LI at word level is a sequence labeling problem where each and every word in a sentence is tagged with one of the languages in the predefined set of languages. For many NLP applications, using code-mixed texts, the first but very crucial preprocessing step will be identifying the languages in a given text. In order to address word level LI in code-mixed Kannada-English (Kn-En) texts, this work presents i) the construction of code-mixed Kn-En dataset called CoLI-Kenglish dataset, ii) code-mixed Kn-En embedding and iii) learning models using Machine Learning (ML), Deep Learning (DL) and Transfer Learning (TL) approaches. Code-mixed Kn-En texts are extracted from Kannada YouTube video comments to construct CoLI-Kenglish dataset and code-mixed Kn-En embedding. The words in CoLI-Kenglish dataset are grouped into six major categories, namely, “Kannada”, “English”, “Mixed-language”, “Name”, “Location” and “Other”. Code-mixed embeddings are used as features by the learning models and are created for each word, by merging the word vectors with sub-words vectors of all the sub-words in each word and character vectors of all the characters in each word. The learning models, namely, CoLI-vectors and CoLI-ngrams based on ML, CoLI-BiLSTM based on DL and CoLI-ULMFiT based on TL approaches are built and evaluated using CoLI-Kenglish dataset. The performances of the learning models illustrated, the superiority of CoLI-ngrams model, compared to other models with a macro average F1-score of 0.64. However, the results of all the learning models were quite competitive with each other.
\end{abstract}


\section{INTRODUCTION} \label{Introduction}

The measure of mineable information is increasing quickly with the rapid growth of social media. In a country like India where multilingualism is popular, people are comfortable in using more than one language and hence usually use a combination of two or more languages to post their comments or messages on social media. However, these comments may be using single script or multiple scripts. The combination of two or more languages in any text is called code-mixing and is gaining popularity among younger generations mainly to use on social media. English is considered as one of the languages for communication in many countries and the keyboard layout of computers and smartphones by default is of Roman script. Even though there are many apps which can be used to write the text in local languages, however, due to technological glitches most of the users prefer Roman script to write the comments in local or code-mixing languages. Analysis of code-mixed text defines a new research trend due to many challenges. As social media content is not governed by the syntax of any of the languages, short sentences are quite common in addition to incomplete sentences and even words. Words may have a high level of typographical errors intentionally holding creative spellings (gr8 for 'great'), phonetic typescript, word play (goooood for 'good'), and abbreviations (OMG for 'Oh my God!'). Generally, the non-English speakers use English words/sentences (through code-mixing and Anglicism) instead of composing online media text using unicode in their languages. They frequently mix multiple languages in comments/messages to express their thoughts on social media making the analysis of code-mixed text an extremely challenging task.
The preliminary step in analyzing code-mixed texts for various applications is identifying the languages used in these texts efficiently as accuracy of the applications depend on the proper identification of languages. Languages may be mixed at paragraph level, sentence level, word level, or even within a word. Despite a lot of work being done in LI, the problem of LI in code-mixed scenario is still a long way from being illuminated~\cite{mandal2018language}. A code-mixed scenario where words of one language are transcribed with words of other languages as prefix or suffix has lot more troubles, particularly due to conflicting phonetics. In such case, proper context can help in tackling issues like ambiguity. However, capturing context in such data is extremely hard. Furthermore, LI faces the problem of accessible code-mixed dataset to build and evaluate the learning models. The bottleneck of data crisis affects the performance of systems quite a lot, generally because of the issue of over-fitting.
India being a multilingual country has a rich heritage of languages and Kannada is one of the Dravidian languages as well as the official language of Karnataka state. People of Karnataka read, write and speak Kannada but many find it difficult to use Kannada script to post messages or comments on social media. While, technological limitations like keyboards of computers and smartphones is one reason, another reason may be the complexity of framing words with consonant conjuncts (vattakshara in Kannada). Hence, most of them use only Roman script or a combination of both Kannada and Roman script to post comments on social media. Kn-En code-mixed text on social media is increasing rapidly. Identifying the language of the words in code-mixed social media text is not only interesting but also challenging. LI at word level, is a sequence labeling problem where each and every word in a sentence is tagged with one of the languages in the predefined set of languages. Sequence labeling problem is a special case of Text Classification (TC). Based on ML, DL and TL, this paper explores Learning Approaches for Code-mixed LI (LA-CoLI) at word level for code-mixed Kn-En text. This study includes:

\begin{itemize}

    \item Developing learning models, namely, CoLI-vectors and CoLI-ngrams based on ML, CoLI-BiLSTM based on DL and CoLI-ULMFiT based on TL approaches,
    \item Developing a code-mixed Kn-En annotated dataset for LI task at word level called as CoLI-Kenglish,
    \item Creating code-mixed Kn-En embeddings for each word by merging word, sub-words and char vectors to build a Skipgram  model which will be used as features in learning models to determine the efficiency of combination of vectors in ML and DL approaches,
    \item Training a general domain Language Model (LM) using raw code-mixed Kn-En texts for ULMFiT model.

\end{itemize}

Comments in Kannada YouTube videos are used to create code-mixed Kn-En annotated dataset, code-mixed Kn-En word embeddings and train the LM. Kn-En annotated dataset and Kn-En word embeddings which will be released on request for research purpose. Overall results illustrate the competitive performance among the learning approaches.

\section{RELATED WORK} 
\label{Related Wor}
In the ongoing history, a lot of works have been explored on code-mixed data of various language pairs for various applications such as LI, Part-of-Speech (POS) tagging etc. Soumil et al.~\cite{mandal2018language} introduced a novel design for LI of code-mixed Bengali-English (Bn-En) and Hindi-English (Hi-En) data using context information. Their dataset consists of 6000 instances each selected from the datasets prepared by Mandal et al.~\cite{mandal2018preparing} and Patra et al.~\cite{patra2018sentiment} for Bn-En and Hi-En language pairs respectively. They performed multichannel neural associations merging CNN and LSTM coupled with BiLSTM-CRF for word-level LI of code-mixed data to achieve 93.28\% and 93.32\% accuracies on the test sets of two language pairs. A novel strategy for incremental POS tagging of code-mixed Spanish/English corpus is proposed by Paul et al.~\cite{rodrigues2013part}. Utilizing dynamic model switching to get an indicator function which emits term-by-term LI tags, their baseline framework obtained an overall accuracy of 77.27\%. The indicator function also regulates the output and picks the most reasonable tagging model to use for a given term. Nguyen et al.\cite{nguyen2013word} introduced experiments on LI of individual words in multilingual conversational data crawled from one of the biggest online networks in Netherlands for Turkish-Dutch speakers during May 2006 to October 2012. Albeit Dutch and Turkish language words rule the discussion, English fixed phrases (e.g. ‘no comment’, ‘come on’) are incidentally observed. They evaluated strategies from different points of view on how language recognizable proof at word level can be utilized to analyze multilingual data. The highly informal spelling in online conversations and the events of named substances was used as test set. For their experiments with multilingual online conversations, they first tag the language of individual words utilizing language models and dictionaries and then incorporate context to improve the performance and achieved an accuracy of 98\%. Results uncover that language models are more robust than dictionaries and adding context improves the performance.
Sarkar et al.~\cite{sarkar2016part} proposed a Hidden Markov Model (HMM) dependent POS tagger for code-mixed Bengali-English (Bn-En), Hindi-English (Hn-En) and Tamil-English (Ta-En) shared task datasets of ICON 2015~\footnote{https://ltrc.iiit.ac.in/icon2015/}. They used information from dictionary based methodologies and some word level features to additionally improve the observation probabilities for prediction. Their framework obtained an average overall accuracy (averaged over all three language sets) of 75.60\% in constrained mode and 70.65\% in unconstrained mode. Yashvardhan et al.~\cite{sharma2020bits2020} presents the methodologies to classify Dravidian code-mixed comments according to their polarity in the evaluation of the track 'Sentiment Analysis for Dravidian Languages in Code-Mixed Text' organized by the Forum of Information Retrieval Evaluation (FIRE) 2020~\footnote{http://fire.irsi.res.in/fire/2020/home}. They trained, validated, and tested the model using the Tamil~\cite{chakravarthi2020corpus} and Malayalam~\cite{chakravarthi2020sentiment} code-mixed datasets provided by the organizers. Tamil code-mixed dataset consists of 11335 comments for the train set, 1260 for the validation set and 3149 comments for testing the model. Malayalam code-mix dataset consists of 4851 comments for training, 541 for validating, and 1348 for testing the model. Using Long Short-Term Memory (LSTM) network alongside language-explicit pre-processing and sub-word level portrayal to catch the assumption of the content, they obtained F1-scores of 0.61 and 0.60 and overall ranks of 5 and 12 for Tamil and Malayalam datasets respectively.

\section{Methodology}
\subsection{Construction of Dataset and Tools}
This section describes the functionality used for data collection, preprocessing, training code-mixed word embeddings and building the first ever code-mixed LM for Kn-En language pairs.
\subsubsection{Data Collection}
Data is the most important part of any study and data for NLP tasks are in form of text and speech. As code-mixed text in Kn-En language pair is required for the proposed work, an efficient module that can scrap data from various sources such as social media platforms, online shopping website, etc. is required. youtube-comment-downloader~\footnote{https://github.com/egbertbouman/youtube-comment-downloader} is modified to download 100000 comments from 373 Kannada YouTube videos which amounts to 72815 sentences after preprocessing. The comments were written only in Kannada or only in English or a combination of Kannada and English and in few cases in other languages namely, Hindi, Telugu and Tamil in addition to Kannada or English or both. However, the script of these comments is either Roman or Kannada or a combination of Roman and Kannada. The workflow of data collection module is shown in Figure 1. Data collection module accepts a list of Kannada YouTube video ids as input, downloads the comments, preprocesses them and provides as output a list of sentences extracted from the comments posted on each video.
\begin{figure}[ht!]
\centering
    \includegraphics[width=0.7\textwidth]{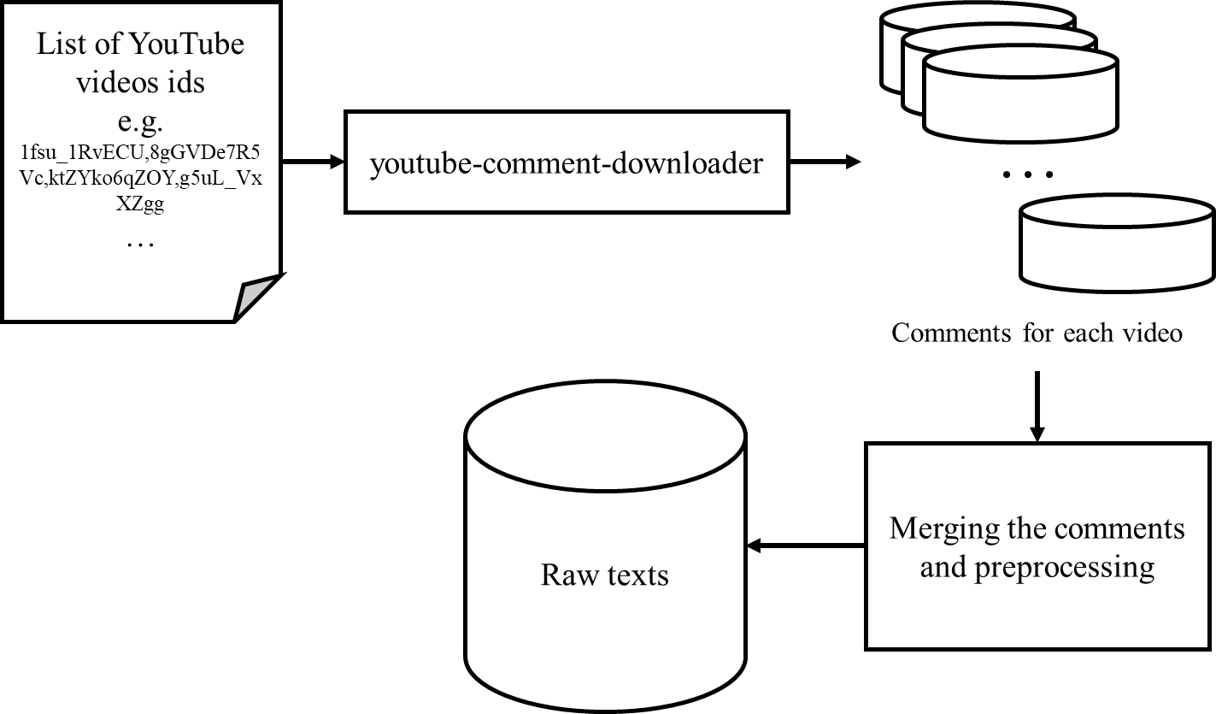}
\caption{Data Collection Module}
\label{fig:ann}
\end{figure}

\subsubsection{Preprocessing}
Comments in social media are unstructured, messy, contain incomplete sentences and words in short forms in addition to code-mixing of two or more languages. All these features increase the complexity of analyzing code-mixed text. Hence, the first step in analyzing these texts is preprocessing, which includes removing duplicate comments, comments in Kannada script, short comments (less than 3 words) and comments consisting of only English words, emojis and unprintable characters. After preprocessing, roughly 90\% of the data is used as raw data to train Kn-En tokenizer, code-mixed Kn-En word embeddings and code-mixed LM for Kn-En language pairs. Remaining 10\% of the data is processed further to create annotated dataset for LI, at the word level. The major problem faced in analyzing code-mixed text is lack of normalization of words.
\subsubsection{Creation of CoLI-Kenglish Dataset}
A small portion (10\%) of the preprocessed code-mixed texts are selected randomly and tokenized into words. These words are tagged manually by two native Kannada speakers (these people are trained about concepts of code-mixed texts and LI task) to generate CoLI-Kenglish dataset. 19432 unique words extracted from nearly 7000 sentences are categorized into 6 classes namely, ‘Kannada’, ‘English’, ‘Mixed-language’, ‘Name’, ‘Location’ and ‘Other’. While the first two classes represent Kannada and English words respectively, ‘Mixed-language’ class represents word created using a combination of Kannada and English in any order. ‘Name’ class represents the names of persons and ‘Location’ class the names of locations or places. Any other words are represented as ‘Other’ class. The words described by ‘Mixed-language’ pose a real challenge to LI task as these words are framed by various combinations of English/Kannada words and Kannada/English affixes (prefixes and suffices). Beauty and also the complexity of these mixed-language words is that the word pattern depends on an individual and users posting comments on social media is increasing day-by-day. Description and samples of tokens are given in Table 1.

\begin{table}[]

    \centering
    \caption{\small Description and samples of tokens in CoLI-Kenglish dataset}
    \scalebox{1}{
    \begin{tabular}{|l|c|l|l|}
    \hline
        \bf Category &\bf Tag&\bf  Description &\bf  Samples \\
        \hline
        Kannada & kn &Kannada words written in Roman script & \begin{tabular}[l]{@{}l@{}} kopista (one who get angry soon),\\ baruthe (will come),\\barbeku (must come) \end{tabular}\\
        \hline
        English & en & Pure English words & small, need, take, important\\
        \hline
        Mixed-language & kn-en & \begin{tabular}[l]{@{}l@{}}Combination of Kannada and English\\words in Roman script \end{tabular}& \begin{tabular}[l]{@{}l@{}} coolagiru (cool + agiru, be cool),\\ leaderge (leader + ge, to a leader),\\ homealli (home + alli, inside home)\end{tabular}\\
        \hline
        Name & name & \begin{tabular}[l]{@{}l@{}} Words that indicate name of person\\(including Indian names)\end{tabular} & Madhuswamy, Hemavati, Swamy\\
        \hline
        Location & location&Words that indicate locations & Karnataka, Tumkur, Bangalore\\
        \hline
        Other &other& \begin{tabular}[l]{@{}l@{}} Words not belonging to\\any of the above categories \\and words of other languages \end{tabular} & \begin{tabular}[l]{@{}l@{}}Znjdjfjbj – not a word\\kannada words in kannada script\\hindi words in Devanagari script\\hindi words in Roman script\\tamil words in Tamil script \end{tabular}\\
        \hline

    \end{tabular}
    
    }
\end{table}

\subsubsection{Word Embeddings}
Word embeddings are seen as the key ingredient for many NLP tasks and has been proved as an efficient representation for characterizing the statistical properties of natural languages~\cite{peters1802deep}. In addition to providing text to numeric vector conversion that is understandable to Neural Networks (NN), they model the complex characteristics of words, such as syntax and semantics which vary across linguistic contexts. Word embeddings consisting of word, sub-words, and char vectors are trained on 90\% of the preprocessed Kn-En code-mixed raw data which is in the form of sentences. The steps to train the vectors as follows:

\begin{itemize}
    \item \textbf{Word vectors: }By tokenizing sentences to words, code-mixed word2vec model of size 200 is trained on the words based on Skipgram model using gensim~\footnote{https://pypi.org/project/gensim/} library
    \item \textbf{Sub-word vectors: }A sub-word is a substring of a word. BPEmb~\footnote{https://nlp.h-its.org/bpemb/} tools are used to split each word to sub-words. Similar to word vectors a code-mixed sub-word2vec of size 100 is trained on the sub-words based on Skipgram model
    \item \textbf{Character vectors: }A char2vec model of size 30 is trained on all characters which appear in the text based on Skipgram model
\end{itemize}

The sizes of the vector's dimensions selected for the proposed word embeddings are set based on the average unique tokens of each type in the dataset (words, sub-words, and characters). A sentence is made up of several words and each word can be decomposed into several sub-words and several characters. Hence, a word vector is extended by sub-words vectors and character vectors. In order to have a fixed length vector representation for words, the number of sub-words is fixed as the maximum of the number of sub-words of all the words in the vocabulary and similarly the number of characters is fixed as the maximum of the number of characters of all the words. For each word, word2vec, sub-word2vec, and char2vec Skipgram based models are trained as mentioned above.
As the number of sub-words is not the same for all words, sub-word2vec of a word is padded with zeros depending on the difference between the maximum of the number of sub-words of all the words and the number of sub-words in a word. Similarly, char2vec is padded with zeros depending on the difference between the maximum of the number of characters of the words and number of characters in a word. Finally, word2vec, sub-word2vec, and char2vec vectors are merged together to obtain one vector for each word as shown in Figure 2. Table 2 gives a glimpse of the size of the all vectors used to obtain a vector for a word.

\subsubsection{Kn-En Tokenizer}
Tokenization is an initial but very crucial step in many token level classification tasks such as POS~\cite{wang2015part}, Named Entity Recognition (NER), and token level LI~\cite{rai2021study}. Many pre-trained tokenizers are available in NLTK~\footnote{https://www.nltk.org/} and iNLTK~\footnote{https://pypi.org/project/inltk/} libraries for tokenizing Indian languages but tokenizers for code-mixed text are rarely found. SentencePiece~\footnote{https://github.com/google/sentencepiece} is an unsupervised text tokenizer that utilizes sub-words units e.g., Byte-Pair-Encoding (BPE)~\cite{sennrich2015neural} and unigrams~\cite{kudo2018subword} with the extension of directly training from raw sentences. A Kn-En code-mixed tokenizer is trained on 90\% of the preprocessed Kn-En code-mixed raw texts with a vocabulary size of 10000 using SentencePiece tools. Figure 3 illustrates the procedure of training Kn-En tokenizer and generating vocabulary.

\begin{table}[ht]
\caption{Glimpse of the all vectors size all to form a vector for a word}
\centering
\begin{tabular}{|l|}
\hline
Size of word2vec = 200                                                                                                                                                                                                                                \\ \hline
Size of sub-word2vec = 100                                                                                                                                                                                                                            \\ \hline
Size of char2vec = 30                                                                                                                                                                                                                                 \\ \hline
Maximum of the number of sub-words of all   words in the vocabulary = 8                                                                                                                                                                               \\ \hline
Maximum of the number of characters of   all the words = 10                                                                                                                                                                                           \\ \hline
\begin{tabular}[c]{@{}l@{}}Total size of word2vec = size of word2vec   + 8 x size of sub-word2vec + 10 size of\\ char2vec = 200 + 8x100 + 10x30 = 1300\end{tabular}                                                                                   \\ \hline
\begin{tabular}[c]{@{}l@{}}If a word ‘w’ has 5 sub-words and 6   characters, then word vector will be a\\ combination of word2vec + 5   sub-word2vec + (8-5) sub-word2vec zero paddings +\\ 6 char2vec + (10-6)   char2vec zero paddings\end{tabular} \\ \hline
\begin{tabular}[c]{@{}l@{}}Sub-word2vec zero paddings will be of the   size of sub-word2vec and char2vec\\ zero paddings will be of the size of   char2vec.\end{tabular}                                                                              \\ \hline
\end{tabular}
\end{table}

\begin{figure}[ht]
\centering
    \includegraphics[width=0.7\textwidth]{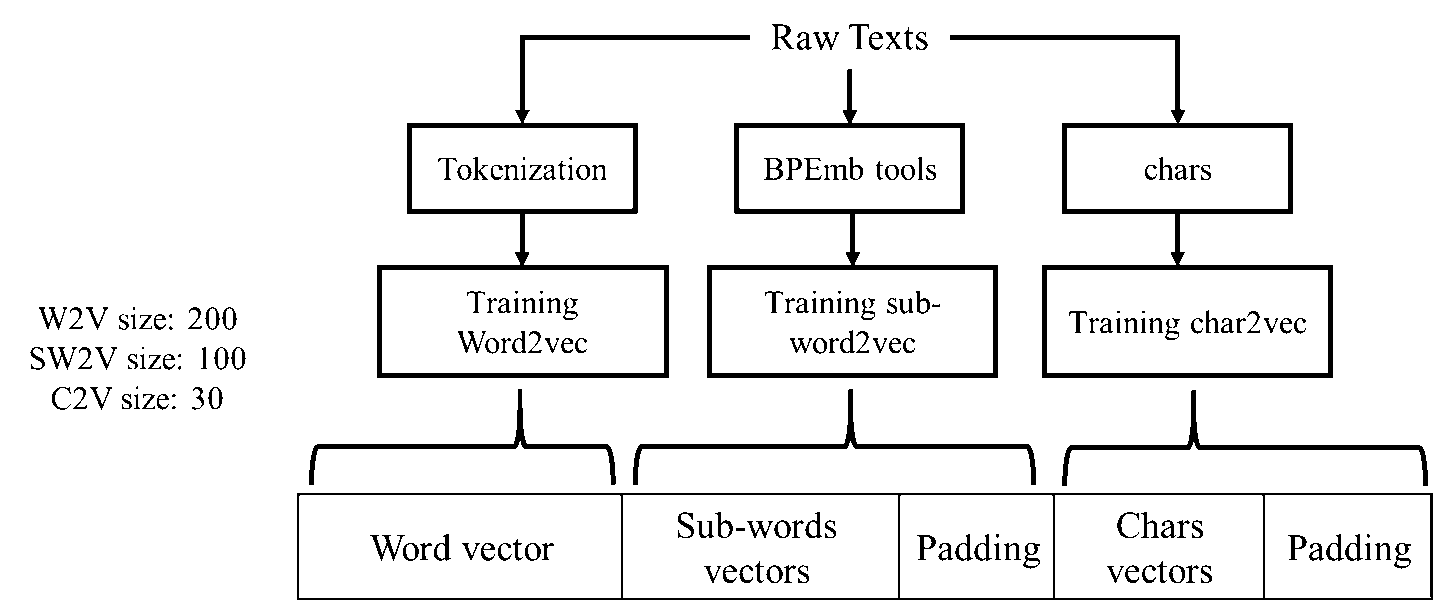}
\caption{Merging word2vec, sub-word2vec, and char2vec vectors}
\label{fig:ann}
\end{figure}

\begin{figure}[ht]
\centering
    \includegraphics[width=0.7\textwidth]{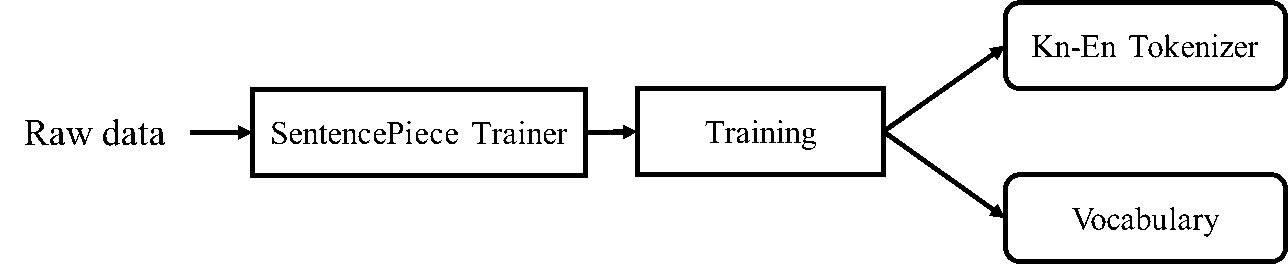}
\caption{Procedure of generating Kn-En tokenizer and Vocabulary}
\label{fig:ann}
\end{figure}

\subsubsection{Language Model}
LM is a probability distribution over the sequence of words, in other words, LM is able to predict next word(s) in a given sequence of words and window~\cite{howard2018universal}. It has applications in NLP tasks such as “Smart Compose” feature in Gmail that suggests next words in sequence. Voice to text conversion, speech recognition, sentiment analysis, text summarization, and spell correction are other NLP tasks where LMs can be used. Further, a LM can be seen as a statistical tool that can learn and analyze the natural languages’ patterns. LM has got more attention with TL where the knowledge of one (source) model is transferred to another (target) model. Raw text collected from YouTube video comments (as mentioned in section 3.1.1) have been used to train a tokenizer using SentencePiece library as explained in section 3.1.5. This is then used along with raw texts to train the LM for Kn-En code-mixed text with a vocabulary size of 10000. Fast.ai~\footnote{https://nlp.fast.ai/} library is used to train the LM for 150 training epochs with various learning rates. More details are given in 3.2.4.

\subsubsection{N-grams Model}
One of the challenges of LI is the structure of words in natural language. For example, it is very common in English to see the letter “q” to be followed by letter 'u' in words such as question, quarrel, qualifications, quietness, etc. However, this rule is not followed in many code-mixed texts. Since one of the primary advantages of character n-grams is language independence~\cite{kruczek2020n} it can be utilized for any language including code-mixed texts to capture the structure of words that has been written in a different script. In this study, a feature engineering module that generates a feature set for a given text is implemented. The feature set consists of prefixes and suffixes of length 1, 2 and 3 along with char ngrams (n = 2, 3, 5) from words, and char ngrams (n = 1, 2, 3) from sub-words.

\subsection{Learning Models}
Four learning models, namely, CoLI-ngrams, CoLI-vectors, CoLI-BiLSTM, and CoLI-ULMFiT are proposed for the Kn-En code-mixed LI task at word level. The learning models based on ML, DL, and TL approaches are constructed and evaluated using CoLI-Kenglish dataset and the tools constructed as mentioned above. All the four learning models are explained below:

\subsubsection{CoLI-ngrams}
This model is an ensemble of three ML classifiers namely, Linear SVC (LSVC), Multi-Layer Perceptron (MLP) and Logistic Regression (LR) with ‘soft’ voting. Values of the parameters used in these classifiers are given in Table 3. Figure 4 presents the structure of CoLI-ngrams model which is fed with count vectors of ngrams obtained from a feature engineering module described in section 3.1.7.
Char ngrams from sub-words are extracted in two steps: i) extracting sub-words from words using BPEmb and ii) generating char ngrams for extracted sub-words. BPEmb provides pre-trained sub-words embeddings for 275 languages that are trained on texts from Wikipedia~\cite{balouchzahi2020mucs}. An embedding with a vocabulary size of 10000 is downloaded for English language to encode and extract sub-words from code-mixing text which helps to extract exact English words from code-mixed words. In code-mixed words, one part of the word may be an English word and rest can be Kannada suffix or prefix or with some characters which do not have any meaning in any language. In other words, sub-Words help in the generation of words that are rarely been seen in training set. Table 4 illustrates the samples of words and corresponding features generated for CoLI-ngrams.

\begin{table}[ht]
\centering
\caption{Parameters for estimators in CoLI-ngrams and CoLI-vectors}
\begin{tabular}{|c|c|}
\hline
\textbf{Estimators} & \textbf{Parameters}                                                                                                                               \\ \hline
Linear SVC          & kernel='linear',probability=True                                                                                                                  \\ \hline
MLP                 & \begin{tabular}[c]{@{}c@{}}hidden\_layer\_sizes=(150,100,50), max\_iter=300,  \\ activation = 'relu', solver='adam', random\_state=1\end{tabular} \\ \hline
LR                  & Default parameters                                                                                                                                \\ \hline
\end{tabular}
\end{table}

\begin{table}[ht]
\centering
\caption{Samples of words and corresponding features generated for CoLI-ngrams model}
\begin{tabular}{|l|l|l|}
\hline
\textbf{Word}                                                                     & \textbf{Language (tag)} & \textbf{Sub-words and ngrams of   sub-words}                                                                                                                                                                        \\ \hline
\begin{tabular}[c]{@{}l@{}}Nayigalige\\    \\ (in English: for dogs)\end{tabular} & Kannada (Kn)            & \begin{tabular}[c]{@{}l@{}}'nayigalige', 'ige', 'nay',   'ge', 'na', 'e', 'n', '\_nay', \\ 'nayi', 'ayig', 'yiga', 'igal', 'gali', 'alig',   'lige', \\ 'ige\_','\_ay\_', '\_ig\_', '\_al\_', '\_ig\_'\end{tabular} \\ \hline
\begin{tabular}[c]{@{}l@{}}Dogsgalige\\    \\ (in English: for dogs)\end{tabular} & Mixed-language (Kn-En)  & \begin{tabular}[c]{@{}l@{}}'dogsgalige', 'ige',   'dog','ge', 'do', 'e', 'd', '\_dog', \\ 'dogs', 'ogsg', 'gsga', 'sgal', 'gali',   'alig', 'lige', \\ 'ige\_', '\_og\_', '\_al\_', '\_ig\_'\end{tabular}           \\ \hline
\end{tabular}
\end{table}

\subsubsection{CoLI-vectors}
This model uses estimators as in CoLI-ngrams model but trained on vectors for words in the training set by utilizing embedding module that generates word embed-dings for words, sub-words and characters from raw text as discussed earlier. The purpose of developing CoLI-vectors model is to compare the performances of voting classifiers with different features and also to compare the efficiency of proposed word embedding architecture using ML and DL approaches. Figure 5 gives the structure of CoLI-vectors model.

\begin{figure}[ht]
\centering
    \includegraphics[width=0.7\textwidth]{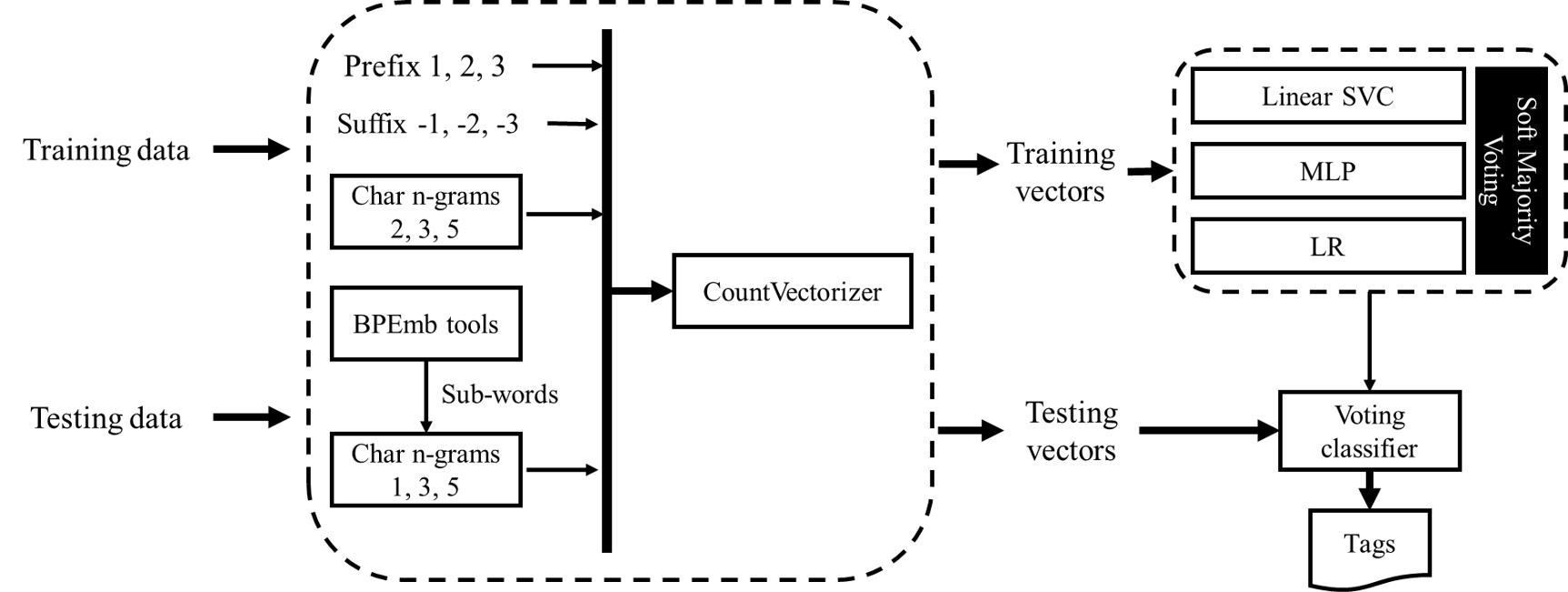}
\caption{Structure of CoLI-ngrams model}
\label{fig:ann}
\end{figure}

\subsubsection{CoLI-BiLSTM}
Learning models based on DL approach have excelled conventional models based on ML approach in various NLP tasks, such as Sentiment Analysis, NER etc.~\cite{minaee2021deep}. CoLI-BiLSTM model is a sequence processing model based on Bidirectional Long Short Term Memory (BiLSTM) architecture. It utilizes the feature vectors obtained from proposed word embedding model. A BiLSTM comprises of two LSTMs that take the input in forward as well as in backward direction. In other words, at every time step BiLSTM networks have both backward and forward information about the sequence~\cite{huang2015bidirectional,hochreiter1997long}. CoLI-BiLSTM model consists of layers summarized in Table 5. It includes input and embedding layers to load training data and weights from word embedding model and a BiLSTM layer followed by time\_distributed layer. The purpose of using time\_distributed layer is to keep one-to-one relations on input and output on RNNs including LSTM and BiLSTM. This scenario is commonly used in NNs in sequence classification tasks such as POS, NER, etc. The structure of CoLI-BiLSTM model is shown in Figure 6.

\begin{table}[]
\centering
\caption{Layers in CoLI-BiLSTM}
\begin{tabular}{|l|l|l|}
\hline
\textbf{Layer (type)}     & \textbf{Output shape} & \textbf{Param \#} \\ \hline
Input   layer             & {[}(none,   1000){]}  & 0                 \\ \hline
Embedding   layer         & (none,   1000, 1000)  & 19162000          \\ \hline
BiLSTM                    & (none,   1000, 600)   & 3122400           \\ \hline
time\_distributed   layer & (none,   1000, 7)     & 4207              \\ \hline
\end{tabular}
\end{table}

\begin{figure}[ht]
\centering
    \includegraphics[width=0.7\textwidth]{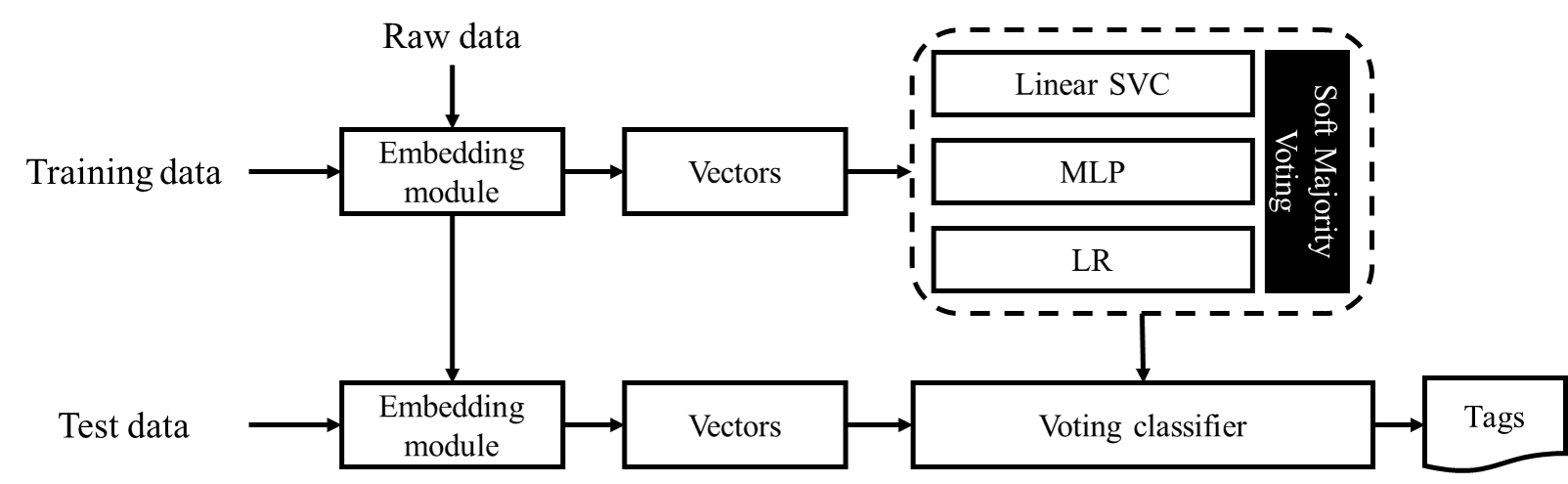}
\caption{Structure of CoLI-vectors model}
\label{fig:ann}
\end{figure}

\begin{figure}[ht]
\centering
    \includegraphics[width=0.7\textwidth]{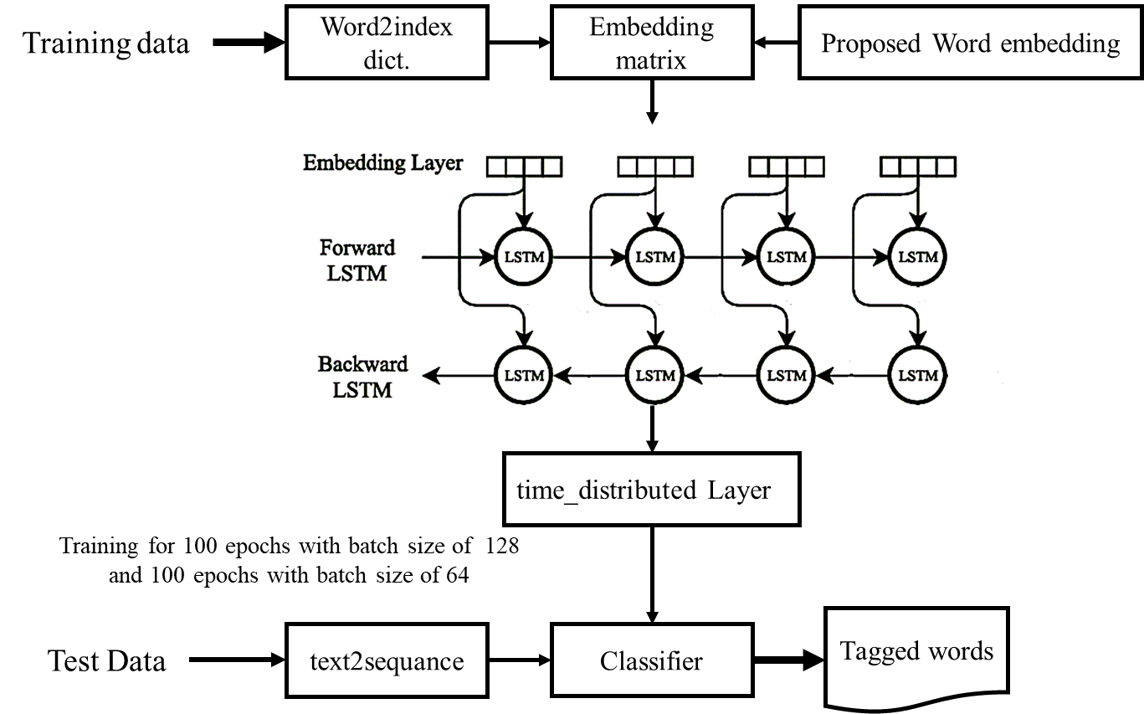}
\caption{Structure of CoLI-BiLSTM model}
\label{fig:ann}
\end{figure}

\subsubsection{CoLI-ULMFiT}
The approach of transferring knowledge of one model called source model to improve the performance of the other model called target model is called TL. Universal Language Model Fine-Tuning (ULMFiT) is one of architectures that utilize the concept of TL~\cite{balouchzahi2020puner}. It consists of training a LM and then transferring the obtained knowledge and fine-tuning the target model with the dataset provided for the given task. Usually in NLP tasks the data used for training LM will be a large corpus with same or different domain from the dataset used for target task. The benefit of training a LM is that once a pre-trained LM is ready, its knowledge can be utilized in different NLP tasks including token level or text level classification, summarization, etc. A pre-trained LM understands the general features of language and then fine-tuning the LM using target task dataset helps in obtaining more properties of specific task. Following the ULMFiT architecture adopted from~\cite{howard2018universal}, CoLI-ULMFiT model includes training a LM from preprocessed code-mixed texts (section 3.1.2), transferring and then fine-tuning the weights using training set (section 3.1.3- CoLI-Kenglish) and finally using weights and knowledge obtained from LM in target LI model. Figure 7 presents the overview of CoLI-ULMFiT model.
Fast.ai library provides necessary modules for the implementation of ULMFiT model. text.models tools from Fast.ai library is used to construct both LM and target LI models. An encoder for Average-Stochastic Gradient Descent (SGD) Weight-Dropped LSTM (AWD-LSTM) implemented using text.models tools consists of a word embedding of size 400, 3 hidden layers and 1150 hidden activations per layer-plugged in with a decoder plus classification layers to create a TC~\cite{merity2017regularizing}.

\begin{figure}[ht]
\centering
    \includegraphics[width=0.7\textwidth]{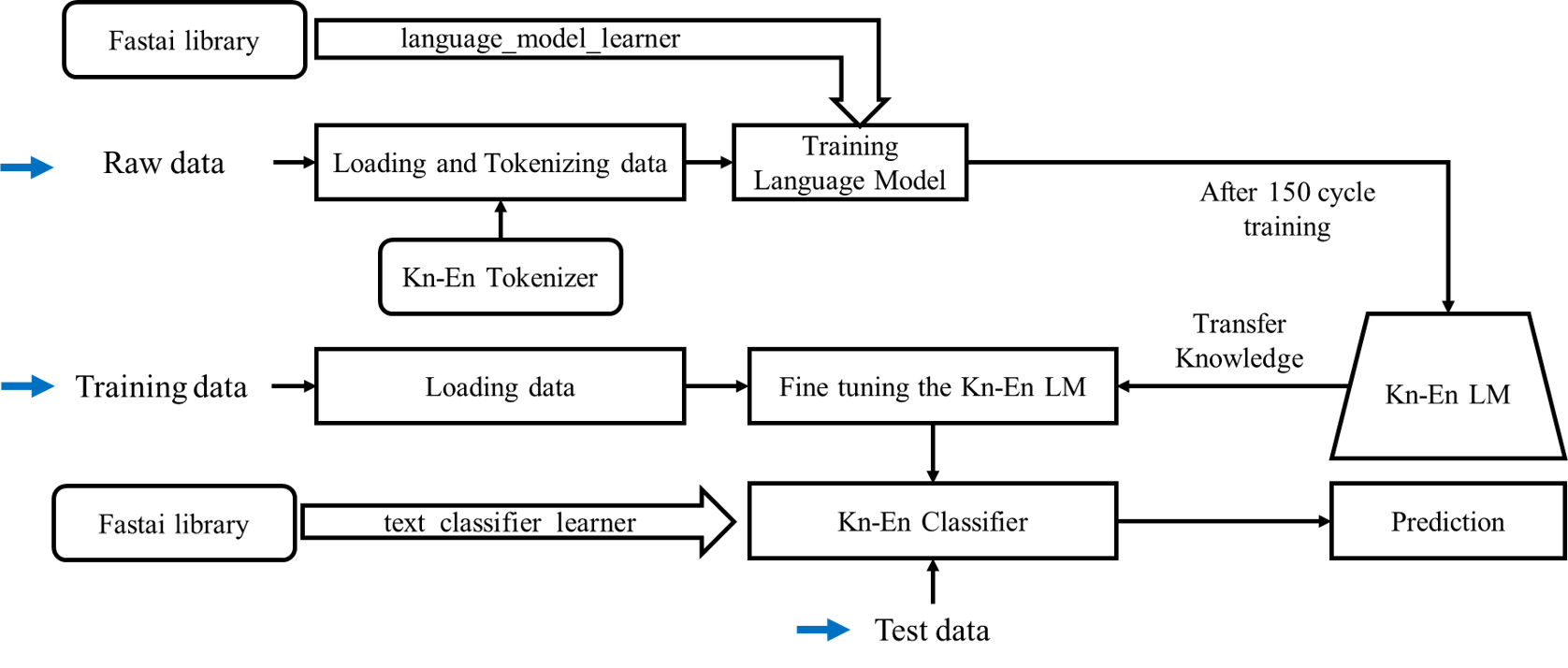}
\caption{Overview of CoLI-ULMFiT model}
\label{fig:ann}
\end{figure}

\section{Experiments and Results}
\subsection{Datasets}
Inspired by~\cite{hande2020kancmd,chakravarthi2020hopeedi,chakravarthi2021findings} in utilization of YouTube code-mixed comments, CoLI-Kenglish dataset has been developed. The construction of the CoLI-Kenglish dataset for LI at word level is mentioned in Section 3.1.3 and the distribution of labels in CoLI-Kenglish dataset is shown in Figure 8. Statistics of raw data and CoLI-Kenglish dataset is summarized in Table 6. Since texts in social media generally do not follow any rules the tagged dataset is highly imbalanced which may result in less F1 score. The dataset also illustrates that nearly 44.8\% words are Kannada words, about 7.5\% words are Kn-En mixed language words like “Dogsgalige” (meaning ‘for dogs’, dogs is an English word and ‘galige’ is a suffix in Kannada) and about 32.32\% words are English words. Approximately, 70\% of the tagged dataset is used for training and remaining 30\% for testing.

\begin{table}[ht]
\centering
\caption{Statistics of datasets}
\begin{tabular}{|l|l|l|l|}
\hline
\textbf{Dataset} & \textbf{Type} & \textbf{No. sentences} & \textbf{No. words} \\ \hline
Raw texts        & unannotated   & 72135                  & 594680             \\ \hline
CoLI-Kenglish DS & annotated     & 700                    & 19432              \\ \hline
\end{tabular}
\end{table}

\begin{figure}[ht]
\centering
    \includegraphics[width=0.7\textwidth]{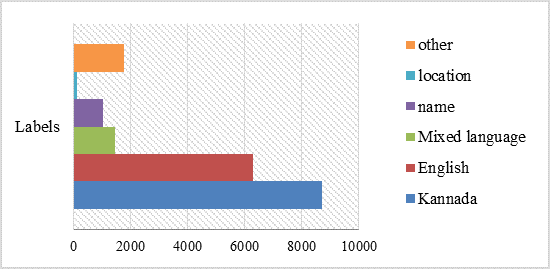}
\caption{Labels distribution over the CoLI-Kenglish DS}
\label{fig:ann}
\end{figure}

\subsection{Results}
This study provides a comparison of the performances of the proposed models for word level LI task in Kn-En code-mixed texts and the results are shown in terms of macro average metrics. Kannada and English are two completely different languages in various terms such as grammar, script, structure, etc., but still performances of models are promising considering the noisiness of the data. However, it is expected in closely related languages, e.g. English-German or Spanish-Italian mixed texts, LI task will be more challenging but availability of more tools for such languages enable models to have more efficient performances.
The performances of the proposed learning models in addition to the performances of individual estimators in case ML models are shown in Table 7. CoLI-Kenglish dataset for word level LI consists of 6 categories and category-wise results in terms of Precision, Recall and F1-score of all the proposed models are shown in Table 8. Further, category-wise comparison of macro average F1-scores of the proposed models is illustrated in Table 9. Results of ML models illustrate that ML classifiers (both individual and ensembled) with character ngrams and affixes outperformed the ML classifiers with proposed word embeddings.

\begin{table}[ht]
\centering
\caption{Comparison of performances among ML models and individual estimators}
\begin{tabular}{|l|l|lll|}
\hline
\multirow{2}{*}{\textbf{Classifier}} & \multirow{2}{*}{\textbf{Features}} & \multicolumn{3}{l|}{\textbf{Performance}}                                                          \\ \cline{3-5} 
                                     &                                    & \multicolumn{1}{l|}{\textbf{Precision}} & \multicolumn{1}{l|}{\textbf{Recall}} & \textbf{F1-score} \\ \hline
Linear SVC                           & word embeddings                    & \multicolumn{1}{l|}{0.35}               & \multicolumn{1}{l|}{0.60}            & 0.37              \\ \hline
LR                                   & word embeddings                    & \multicolumn{1}{l|}{0.37}               & \multicolumn{1}{l|}{0.69}            & 0.40              \\ \hline
MLP                                  & word embeddings                    & \multicolumn{1}{l|}{0.37}               & \multicolumn{1}{l|}{0.64}            & 0.39              \\ \hline
CoLI-vectors                         & word embeddings                    & \multicolumn{1}{l|}{0.36}               & \multicolumn{1}{l|}{0.69}            & 0.39              \\ \hline
Linear SVC                           & ngrams + affixes                   & \multicolumn{1}{l|}{0.73}               & \multicolumn{1}{l|}{0.57}            & 0.62              \\ \hline
LR                                   & ngrams + affixes                   & \multicolumn{1}{l|}{0.74}               & \multicolumn{1}{l|}{0.55}            & 0.60              \\ \hline
MLP                                  & ngrams + affixes                   & \multicolumn{1}{l|}{0.70}               & \multicolumn{1}{l|}{0.60}            & 0.63              \\ \hline
\bf CoLI- ngrams                         & \bf ngrams + affixes                   & \multicolumn{1}{l|}{\bf 0.73}               & \multicolumn{1}{l|}{\bf 0.60}            & \bf 0.64              \\ \hline
CoLI-BiLSTM                          & Proposed vectors                   & \multicolumn{1}{l|}{0.61}               & \multicolumn{1}{l|}{0.74}            & 0.63              \\ \hline
CoLI-ULMFiT                          & A code-mixed LM                    & \multicolumn{1}{l|}{0.42}               & \multicolumn{1}{l|}{0.42}            & 0.41              \\ \hline
\end{tabular}
\end{table}

CoLI-BiLSTM model has been trained for 200 epochs (100 epochs with batch size of 128 and 100 epochs with batch size of 64. The results illustrate that that CoLI-ngrams model based on ML approach trained on a subset of morphological features including char ngrams from words and sub-words along with affixes beats the other models.  CoLI-ULMFiT model has obtained relatively good performance except for words belonging to “location” class which is due to lack of sufficient samples in tagged dataset. Training an LM for ULMFiT architecture efficiently requires very huge dataset.

\begin{table}[ht]
\centering
\caption{Category-wise results of the proposed models}
\begin{tabular}{|c|c|cccccc|}
\hline
\multirow{2}{*}{\textbf{Model}} & \multirow{2}{*}{\textbf{Metric}} & \multicolumn{6}{c|}{\textbf{Labels}}                                                                                                                                                                                        \\ \cline{3-8} 
                                &                                  & \multicolumn{1}{c|}{\textbf{English}} & \multicolumn{1}{c|}{\textbf{Kannada}} & \multicolumn{1}{c|}{\textbf{Mixed-language}} & \multicolumn{1}{c|}{\textbf{Name}} & \multicolumn{1}{c|}{\textbf{location}} & \textbf{other} \\ \hline
\multirow{3}{*}{CoLI-vectors}   & Precision                        & \multicolumn{1}{c|}{0.66}             & \multicolumn{1}{c|}{0.94}             & \multicolumn{1}{c|}{0.02}                    & \multicolumn{1}{c|}{0.29}          & \multicolumn{1}{c|}{0.13}              & 0.14           \\ \cline{2-8} 
                                & Recall                           & \multicolumn{1}{c|}{0.87}             & \multicolumn{1}{c|}{0.60}             & \multicolumn{1}{c|}{0.77}                    & \multicolumn{1}{c|}{0.56}          & \multicolumn{1}{c|}{0.86}              & 0.50           \\ \cline{2-8} 
                                & F1-score                         & \multicolumn{1}{c|}{0.75}             & \multicolumn{1}{c|}{0.74}             & \multicolumn{1}{c|}{0.05}                    & \multicolumn{1}{c|}{0.38}          & \multicolumn{1}{c|}{0.22}              & 0.22           \\ \hline
\multirow{3}{*}{CoLI- ngrams}   & Precision                        & \multicolumn{1}{c|}{0.83}             & \multicolumn{1}{c|}{0.82}             & \multicolumn{1}{c|}{0.87}                    & \multicolumn{1}{c|}{0.56}          & \multicolumn{1}{c|}{0.75}              & 0.57           \\ \cline{2-8} 
                                & Recall                           & \multicolumn{1}{c|}{0.87}             & \multicolumn{1}{c|}{0.89}             & \multicolumn{1}{c|}{0.66}                    & \multicolumn{1}{c|}{0.45}          & \multicolumn{1}{c|}{0.27}              & 0.44           \\ \cline{2-8} 
                                & F1-score                         & \multicolumn{1}{c|}{\textbf{0.85}}    & \multicolumn{1}{c|}{\textbf{0.85}}    & \multicolumn{1}{c|}{0.75}                    & \multicolumn{1}{c|}{\textbf{0.50}} & \multicolumn{1}{c|}{\textbf{0.39}}     & \textbf{0.50}  \\ \hline
\multirow{3}{*}{CoLI-BiLSTM}    & Precision                        & \multicolumn{1}{c|}{0.74}             & \multicolumn{1}{c|}{0.87}             & \multicolumn{1}{c|}{0.69}                    & \multicolumn{1}{c|}{0.21}          & \multicolumn{1}{c|}{0.14}              & 1.00           \\ \cline{2-8} 
                                & Recall                           & \multicolumn{1}{c|}{0.75}             & \multicolumn{1}{c|}{0.70}             & \multicolumn{1}{c|}{\textbf{0.87}}           & \multicolumn{1}{c|}{0.40}          & \multicolumn{1}{c|}{0.71}              & 1.00           \\ \cline{2-8} 
                                & F1-score                         & \multicolumn{1}{c|}{0.74}             & \multicolumn{1}{c|}{0.78}             & \multicolumn{1}{c|}{0.77}                    & \multicolumn{1}{c|}{0.27}          & \multicolumn{1}{c|}{0.24}              & 1.00           \\ \hline
\multirow{3}{*}{CoLI-ULMFiT}    & Precision                        & \multicolumn{1}{c|}{0.68}             & \multicolumn{1}{c|}{0.71}             & \multicolumn{1}{c|}{0.68}                    & \multicolumn{1}{c|}{0.09}          & \multicolumn{1}{c|}{0.0}               & 0.34           \\ \cline{2-8} 
                                & Recall                           & \multicolumn{1}{c|}{0.81}             & \multicolumn{1}{c|}{0.71}             & \multicolumn{1}{c|}{0.67}                    & \multicolumn{1}{c|}{0.03}          & \multicolumn{1}{c|}{0.0}               & 0.30           \\ \cline{2-8} 
                                & F1-score                         & \multicolumn{1}{c|}{0.74}             & \multicolumn{1}{c|}{0.71}             & \multicolumn{1}{c|}{0.67}                    & \multicolumn{1}{c|}{0.04}          & \multicolumn{1}{c|}{0.0}               & 0.32           \\ \hline
\end{tabular}
\end{table}

\begin{table}[]
\centering
\caption{Category-wise comparison of F1-score of the proposed models}
\begin{tabular}{|c|c|c|c|c|c|c|}
\hline
\textbf{Model} & \textbf{English} & \textbf{Kannada} & \textbf{Mixed-language} & \textbf{Name} & \textbf{location} & \textbf{other} \\ \hline
CoLI-vectors   & 0.75             & 0.74             & 0.05                    & 0.38          & 0.22              & 0.22           \\ \hline
CoLI- ngrams   & 0.85             & 0.85             & 0.75                    & 0.5           & 0.39              & 0.5            \\ \hline
CoLI-BiLSTM    & 0.74             & 0.78             & 0.77                    & 0.27          & 0.24              & 1              \\ \hline
CoLI-ULMFiT    & 0.74             & 0.71             & 0.67                    & 0.04          & 0                 & 0.32           \\ \hline
\end{tabular}
\end{table}

\section{Conclusions}
This study explores four learning models, the CoLI-ngrams, the CoLI-vectors, the CoLI-BiLSTM and the CoLI-ULMFiT for Kn-En code-mixed LI at the word level. While CoLI-ngrams and CoLI-vectors are based on ML approaches, CoLI-BiLSTM and CoLI-ULMFiT are based on DL and TL approaches respectively. Due to lack of Kn-En code-mixed dataset at word level for LI and tools to process Kn-En code-mixed data, comments in Kannada YouTube videos were scrapped and processed to construct CoLI-Kenglish tagged dataset, Kn-En code-mixed word embeddings and Kn-En code-mixed LM. CoLI-Kenglish dataset was manually tagged by Kannada speakers and grouped into six categories. Kn-En code-mixed word embeddings was constructed by merging word, sub-words, and characters vectors which were built using Skipgram model. This embedding is used as features in CoLI-vectors and CoLI-BiLSTM models and a subset of morphological characteristics are used as features in CoLI-ngrams model.
CoLI-ULMFiT utilizes ULMFiT architecture to transfer the knowledge of a pre-trained Kn-En LM to a LI model. The results of the proposed models illustrate that CoLI-ngrams utilizing morphological features outperformed all other models with an average macro F1-score of 0.64. Further, CoLI-ULMFiT model also obtained similar overall performance, except for the “location” category. The results obtained by CoLI-vectors and CoLI-BiLSTM models illustrate the superiority of DL approach over ML approach in using proposed embedding.
As the generated Kn-En code-mixed LM and Kn-En code-mixed word embedding can be used for other Ka-En code-mixed NLP tasks they will be released publicly along with CoLI-Kenglish dataset. In the future, it is planned to enrich both unannotated and annotated dataset, construct a balanced label distribution and to explore different feature sets and models based on different learning approaches. It is also planned, to bring morphological features into the vector space, to determine the enhancement potential of the proposed models.

\section*{ACKNOWLEDGMENTS}

The work was supported by the Mexican Government through the grant A1-S-47854 of the CONACYT, Mexico and grants 20211784, 20211884, and 20211178 of the Secretaría de Investigación y Posgrado of the Instituto Politécnico Nacional, Mexico.

\bibliographystyle{unsrt}  
\bibliography{references}

\end{document}